\documentclass[conference]{IEEEtran}
\IEEEoverridecommandlockouts
\usepackage{cite}
\usepackage{amsmath,amssymb,amsfonts}
\usepackage{algorithmic}
\usepackage{graphicx}
\usepackage{booktabs}
\usepackage{subfig}
\usepackage{textcomp}
\usepackage[table,xcdraw]{xcolor}
\usepackage{listings}
\usepackage{microtype}
\usepackage{comment}
\usepackage{url}
\usepackage{siunitx}

\definecolor{dkgreen}{rgb}{0,0.6,0}
\definecolor{gray}{rgb}{0.5,0.5,0.5}
\definecolor{mauve}{rgb}{0.58,0,0.82}

\makeatletter
\newcommand{\linebreakand}{%
  \end{@IEEEauthorhalign}
  \hfill\mbox{}\par
  \mbox{}\hfill\begin{@IEEEauthorhalign}
}
\makeatother

\usepackage[ruled, vlined, linesnumbered]{algorithm2e} 
\RestyleAlgo{ruled}
\SetKwProg{Init}{\texttt{setupImpl}}{}{}
\SetKwProg{Step}{\texttt{stepImpl}}{}{}
\SetKwProg{getFeetH}{\texttt{Robot.get\_feet\_H}}{}{}
\SetKwProg{getWorldTransform}{\texttt{wrappers.getWorldTransform}}{}{}
\SetKwProg{getWorldTransformNew}{\texttt{SFwrappers.getWorldTransform}}{}{}

\def\BibTeX{{\rm B\kern-.05em{\sc i\kern-.025em b}\kern-.08em
    T\kern-.1667em\lower.7ex\hbox{E}\kern-.125emX}}
\begin{document}

\title{A Flexible MATLAB/Simulink Simulator for Robotic Floating-base Systems in Contact with the Ground
}

\author{\IEEEauthorblockN{Nuno Guedelha\IEEEauthorrefmark{1}\IEEEauthorrefmark{2},
Venus Pasandi\IEEEauthorrefmark{1}\IEEEauthorrefmark{2},
Giuseppe L'Erario\IEEEauthorrefmark{1}, 
Silvio Traversaro\IEEEauthorrefmark{1}, and
Daniele Pucci\IEEEauthorrefmark{1}}
\IEEEauthorblockA{\IEEEauthorrefmark{1}
\textit{Artificial and Mechanical Intelligence}, 
\textit{Istituto Italiano di Tecnologia}, 
Genova, Italy,
firstname.lastname@iit.it}
\IEEEauthorrefmark{2}
These two authors contributed equally to this work.
}

\IEEEoverridecommandlockouts
\IEEEpubid{\makebox[\columnwidth+\columnsep+\columnwidth]{\parbox[t]{\textwidth}{\copyright2022 IEEE. Personal use of this material is permitted. Permission from IEEE must be obtained for all other uses, in any current or future media, including reprinting/republishing this material for advertising or promotional purposes, creating new collective works, for resale or redistribution to servers or lists, or reuse of any copyrighted component of this work in other works. \hfill} \hfill}}

\maketitle

\begin{abstract}
Physics simulators are widely used in robotics fields, from mechanical design to dynamic simulation, and controller design.
This paper presents an open-source MATLAB/Simulink simulator for rigid-body articulated systems, including manipulators and floating-base robots.
Thanks to MATLAB/Simulink features like MATLAB system classes and Simulink function blocks, the presented simulator combines a programmatic and block-based approach, resulting in a flexible design in the sense that different parts, including its physics engine, robot-ground interaction model, and state evolution algorithm are simply accessible and editable.
Moreover, through the use of Simulink dynamic mask blocks, the proposed simulation framework supports robot models integrating open-chain and closed-chain kinematics with any desired number of links interacting with the ground.
The simulator can also integrate second-order actuator dynamics.
Furthermore, the simulator benefits from a one-line installation and an easy-to-use Simulink interface.
\end{abstract}


\section{Introduction}

Physics simulators provide a rapid, inexpensive and safe test platform for validating theoretical investigations and improving robots prototype (e.g., robot mechanical and control design).
In this paper, we present an open-source MATLAB/Simulink library for the simulation of robots with rigid bodies, including manipulators and floating-base robots.

Robotic simulators include fundamental features like 
(i) physics engine for dynamic modelling, and
(ii) contact detection and model,
and some functionality features like motion visualization, importing scenes and meshes, low reality gap, low complexity, high reproducibility, simple scenario and environment construction, low resource cost, automation features (i.e., allows automated testing and headless, scripted, or parallel execution), high reliability (i.e., simulator stability, timing, and synchronization), and high interface stability~\cite{afzal2020study}.

Various simulators have been proposed for robotic systems including commercial and open-source simulators \cite{collins2021review}.
Commercial simulators often support multiple physics engines.
Apart from the possibility to switch between multiple physics engines, simulators rarely support the possibility of editing their physics engine.
From the contact model perspective, simulators use a variety of models.
Different contact models have various parameters affecting their reality gap and stability.
Usually, in commercial simulators, the contact model cannot be modified by the user, while some of its parameters could be adjusted by the user considering the desired robot model, environment, and motion scenario.
Instead, in open-source simulators, the contact model is mostly available in detail, and also the user has the possibility to modify this model by considering the simulation test features~\cite{acosta2022validating}.
Considering the above explanation and the fact that commercial simulators have high resource costs, open-source simulators are an interesting choice for robotic developers \cite{afzal2020study}.

Robotics researchers and engineers widely use MATLAB and Simulink for the design, simulation, and verification of robotic systems.
In this regard, some open-source robotic simulators based on MATLAB have been presented in the literature, including OpenMAS~\cite{openmas2020} and FROST~\cite{Hereid2017FROST}.
However, OpenMAS is an agent-oriented, and not a physical, simulator with support for human-environment contacts.
FROST is based on both MATLAB and Mathematica.
To the best of the author's knowledge, an open-source MATLAB/Simulink library for simulating floating base robotic systems in contact with the environment is still missing.

\IEEEpubidadjcol

In this paper, we present an open-source flexible robotic simulator with an easy-to-use interface for rigid body robots including manipulators and free-floating robots.
The proposed simulator is a Simulink library that relies on iDynTree which is an open-source multibody dynamics library developed in C++ for free-floating robots \cite{nori2015icub}.
We used iDynTree functionalities wrapped by Whole-Body toolbox \cite{FerigoControllers2020} through Simulink Function blocks which improve the whole model execution speed and provide the implementation flexibility to replace iDynTree with the desired physics engine.
The proposed simulator is designed and implemented by the use of MATLAB System classes and the programmatic call to dynamics computation functions.
As a result, details of the proposed simulator logic including the contact detection and model, and the state evolution algorithm is simply accessible and modifiable.
Moreover, thanks to the Dynamic Mask Subsystem feature in Simulink, the proposed simulator can be used for various robot models including robots with open-chain or closed-chain kinematics with any desired number of links interacting with the ground.
Finally, a visualizer that relies on iDynTree is also provided for the visualization of robot motion.
Compared to the existing open-source robotic simulators based on MATLAB, our proposed simulator has the following advantages:
(i) Simulink easy-to-use interface,
(ii) implementation flexibility allowing editing the physics engine, contact detection and model, and the whole logic of the proposed simulator,
(iii) capability to simply be integrated into the sensor and motor dynamics,

The rest of the paper is organized as follows.
Section~\ref{sec:methods} describes the architecture of the proposed Simulink library and its key elements.
Various features of the proposed library as well as the performance are validated in Section~\ref{sec:results}.
Finally, Section~\ref{sec:conclusion} draws the conclusions.
\graphicspath{{./figs/Methods/}}

\section{Methods}\label{sec:methods}

\subsection{Overview of the Simulator}\label{sec:methods:overview-of-the-simulator}

\begin{figure}
    \centering
        \includegraphics[width=0.45\textwidth]{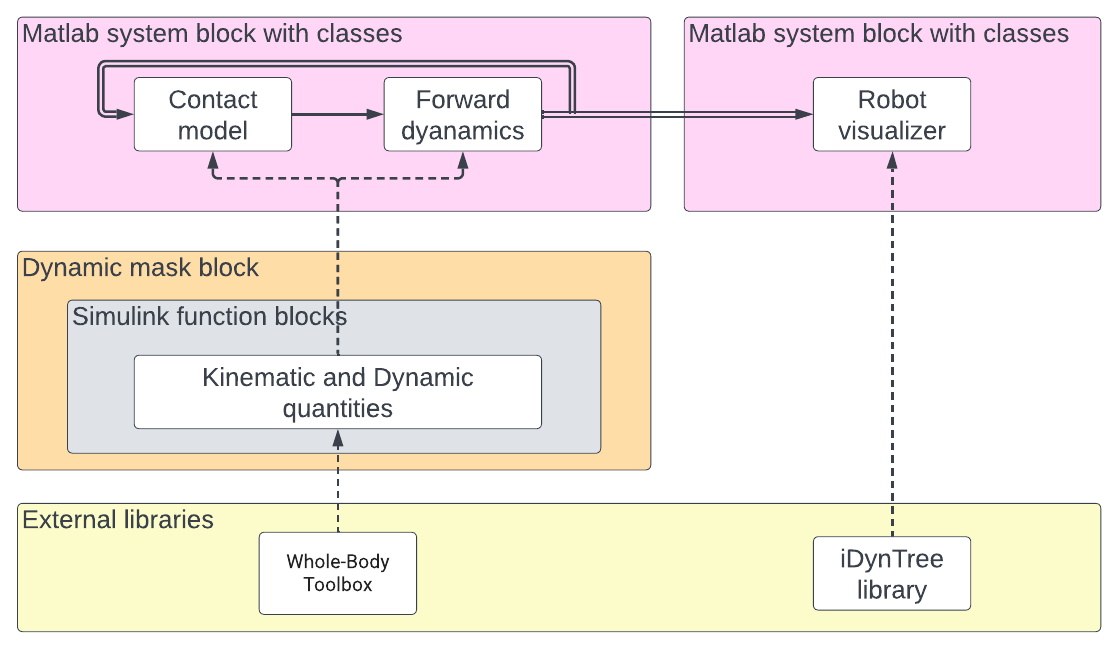}
    \caption{Overview of the Simulator. Solid, single and double arrows represent the data flow between the different blocks, double arrows being used for buses. Dashed arrows denote the programmatic call of its source block.}
    \label{fig:simulator-overview}
\end{figure}

As shown in Figure \ref{fig:simulator-overview}, the Simulator core consists of three main blocks, namely two MATLAB System blocks and a set of Simulink Function blocks redefined as Dynamic Mask Subsystems.
The first MATLAB System Block contains a MATLAB class for computing the time evolution of the robot states.
For this purpose, the contact model computes the reaction forces applied to the robot by the ground, and a forward dynamic model computes the robot acceleration considering the applied forces.
For both contact and forward dynamic models, we need some kinematic and dynamic quantities of the robot, like link Jacobian matrices, inertia matrix, Coriolis and Centrifugal effects vector, etc.
For computing the kinematic and dynamic quantities, we use Simulink Functions which call different functionalities of the Whole-Body Toolbox~\cite{FerigoControllers2020}.
The second MATLAB System block contains a MATLAB class for visualizing the robot's motion.
For the visualization, we use the iDynTree library~\cite{nori2015icub}.

\subsection{MATLAB System Block with Classes}\label{sec:methods:matlab-system-block-with-classes}

MATLAB System blocks allow the object-oriented implementation of algorithms and their use in Simulink~\cite{MatlabSystemBlockWebPage}.
They can be used to simulate a dynamical system's state evolution with time-varying inputs. Two main methods of this block are the \texttt{setupImpl} method, in which attributes are initialized, and classes are instantiated, and the \texttt{stepImpl}, in which the algorithm is implemented
(the same logic applies to the \texttt{Robot visualizer}).

The behavior of the MATLAB System block simulating the robot state evolution (top left MATLAB System block in Figure \ref{fig:simulator-overview}) depends on the current input and its state (the joint torques and the robot's configuration and velocity, respectively) stored in the internal attributes of the class. We implement this behavior through three pure MATLAB classes: the class \textbf{Robot} computes the forward dynamics of the robot, namely the configuration acceleration driven by the joint torques; the class \textbf{Contact} computes the reaction forces; the class \textbf{State} computes the system's state evolution. The robot acceleration is integrated over the integration step.

\subsection{Simulink Function Block}\label{sec:methods:simulink-function-block}

The design and implementation of complex algorithms were made easier by the use of MATLAB System classes and the programmatic call, through binding MEX functions, to external dynamics computation libraries like iDynTree.
Such an improvement came with a downside, as it significantly slowed down the simulation, lowering the real-time factor. This was due to the model being simulated via the MATLAB interpreter engine which is quite slow compared to Simulink's code generation\footnote{https://www.mathworks.com/help/simulink/ug/simulation-modes.html}. The latter supports only a subset of MATLAB functions and is not compatible with direct calls to MEX (Matlab EXecutable) functions\footnote{\url{https://www.mathworks.com/help/matlab/call-mex-functions.html}}. Simulink Functions allow replacing the former bindings with programmatic calls to custom, self-contained Simulink blocks, like the WBT (Whole-Body Toolbox) blocks which implement a C++ abstraction layer wrapping the iDynTree library through the \textit{BlockFactory} framework~\cite{FerigoControllers2020}. The WBT blocks are code generation ready, allowing a significant execution speed improvement (refer to Section \ref{sec:results}), while keeping a compact and flexible implementation of complex algorithms, and a flexible analysis capability as it allows users to use Simulink's rapid prototyping and visual data analysis abilities.

\subsubsection*{Integrating a Simulink Function example \texttt{simFunc\_qpOASES}}

The Simulink Function \texttt{simFunc\_qpOASES} wraps a WBT block implementing a QP (Quadratic Programming) solver from the qpOASES library\footnote{https://github.com/coin-or/qpOASES}, and is triggered by a MATLAB function call, which prototype matches the WBT block interface, from the class \texttt{Contact} implementing the contact model algorithm.

This model scales to the full set of Simulink Functions, configured as Dynamic Mask Subsystems (refer to Section \ref{sec:methods:dynamic-mask-block}), which implement the kinematics and dynamics computations invoked by the contact model and forward dynamics algorithms.

\subsection{Output Bus with Multiple Numerical Signals}\label{sec:methods:output-bus-with-multiple-numerical-sugnals}

\subsubsection{Motivation} We can wrap the output signals of a MATLAB System block in a bus port~\cite{UsingBusesWithMatlabSystemBlocksWebPage} by setting an interface bus programmatically~\cite{SimulinkBusSpecifyPropertiesOfBusesWebPage}, thus avoiding the usage of additional Simulink "Signal Routing" elements like a "Bus Creator". The same can be done on input signals, in order to:
reduce the clutter in a model growing in complexity;
directly interconnect multiple MATLAB System blocks and move them around without caring about re-arranging signal connection lines.

\subsubsection{Implementation} We integrated such a bus output port in the first MATLAB System block (introduced in \ref{sec:methods:overview-of-the-simulator}) implemented by the binded class \texttt{step\_block} which computes the system state evolution algorithm.
The \texttt{kynDynOut} bus wraps all the kinetic and dynamic quantities computed by that algorithm at each simulation time step: the robot state and Jacobian matrices; the inertia matrix; the generalized bias forces and so on.

\subsection{Dynamic Mask Subsystem}\label{sec:methods:dynamic-mask-block}

Various robot models have different kinematic and dynamic characteristics, like a different number of feet\footnote{we define feet as the links that interact with ground}, and open-chain and closed-chain kinematics with a different number of closed-chains.
The simulation of these different features requires different algorithms and information.
To provide those, we sometimes need to modify the content of some subsystem blocks in the Simulink model, typically for adapting that content to the desired robot model.
That can be simply done by redefining the subsystem block as a Dynamic Mask Subsystem~\cite{MatlabDynamicMaskedSubsystemWebPage}.

\graphicspath{{./figs/Results/}}

\section{Results}\label{sec:results}

In this section we first analyze how swapping the bindings with the Simulink Functions improved the simulation speed, almost reaching a sim-to-real time rate of one. We then compare that same performance with another simulator commonly used in the robotics community, Gazebo, while running a benchmark test we define for that purpose. Finally, we illustrate the flexibility of our simulator by analyzing the process of adding a new feature to the simulation framework, namely the support of a contact model for spherical feet.
All results shall be made available on GitHub\footnote{\url{https://github.com/ami-iit/paper_guedelha_2022_irc_flexible-matlab-simulink-robots-simulator}}. The simulator packages, as well as their dependencies, can be installed through Conda package manager, as described in the project GitHub repository respective section\footnote{\url{https://github.com/ami-iit/matlab-whole-body-simulator#readme}}.

\subsection{Functions targeted for the performance analysis}

In view of evaluating the execution speed improvement resulting from the introduction of Simulink Functions, we considered the minimal required code refactoring, i.e. all the changes done in the wrapper functions implementation with the function prototype and the calling code staying unchanged.

we selected a few critical functions in the class \texttt{Robot} which compute the mass matrix, the generalized bias forces, the feet frames Jacobians, the feet bias accelerations and the world to feet frames transformations.
For each of these methods, the implementation using bindings to the iDynTree library has a significant overhead processing due to conversion of vectors and matrices to a MATLAB format (\texttt{iDynTree.toMatlab()}). In some cases like the world to feet frames transformations, an additional step is required, \texttt{iDynTree.asHomogeneousTransform()} for converting the rotation and translation into a single Homogeneous transformation. When switching to Simulink Functions, that processing is directly integrated in a monolithic Simulink block, which can be called programmatically from the MATLAB System classes.

\subsection{Profiling Methodology}

For characterizing the simulation performance and showcasing the improvements brought by the Simulink Functions, we've used two profilers: the MATLAB profiler\footnote{https://www.mathworks.com/help/matlab/ref/profile.html} which tracks the execution of MATLAB interpreted code from MATLAB scripts, from functions called by user-defined MATLAB Function blocks or classes instantiated by MATLAB System blocks; the Simulink profiler\footnote{https://www.mathworks.com/help/simulink/slref/simulinkprofiler.html} which tracks the execution of native Simulink blocks, MATLAB System blocks themselves, or user-defined (Block Factory, Whole-Body Toolbox) Simulink blocks.

\begin{figure*}
    \centering
    \includegraphics[width=\textwidth]{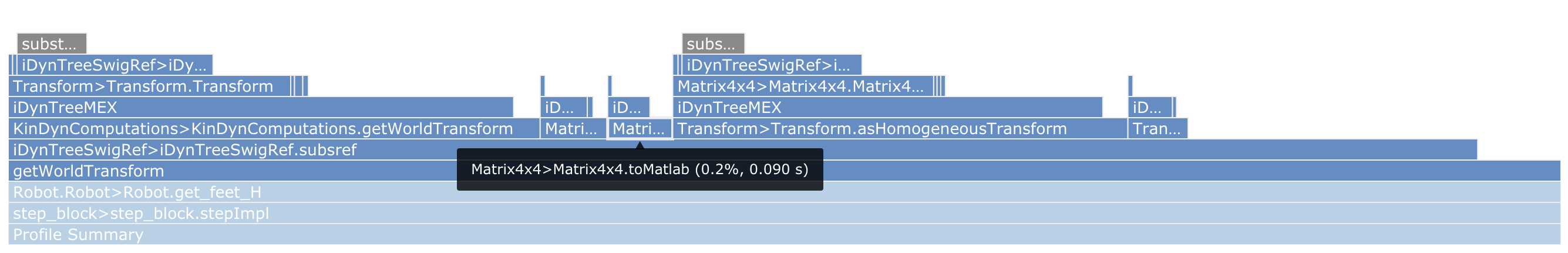}
    \caption{Flame graph generated by the MATLAB interpreter, displaying the MATLAB functions and bindings processing times when running the non-optimised simulator. We can see the proportions between $\texttt{Transform.asHomogeneousTransform}$ and $\texttt{Matrix4x4.toMatlab}$ conversion functions and $\texttt{KinDynComputations.getWorldTransform}$.}
    \label{fig:getFeetHflameGraph}
\end{figure*}

\begin{figure}[t!]
    \centering
    \includegraphics[width=\columnwidth]{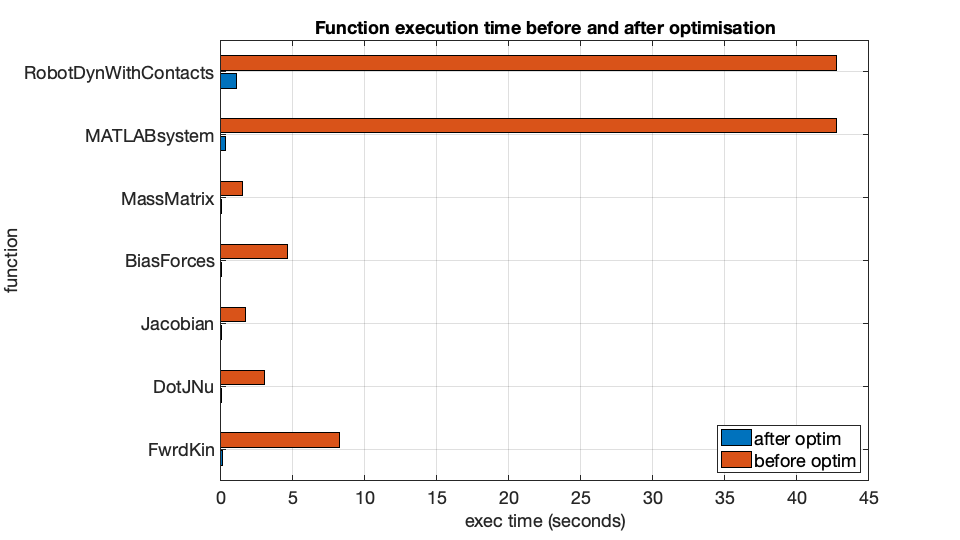}
    \caption{Critical $\texttt{Robot}$ class methods execution time in one second of total simulated time: comparison before and after the optimisation introducing the Simulink Functions.}
    \label{fig:beforeAfterOptimComparison}
\end{figure}

The generated profiler reports provide the number of calls, the self-time (excluding child functions execution time), and the total time. The profile data can be imported to the workspace from which we generate a bar graph (Figure \ref{fig:beforeAfterOptimComparison}) that plots, for each selected method in the class \texttt{Robot}, the total processing time before and after switching to Simulink functions. In Figure \ref{fig:getFeetHflameGraph}, a flame graph shows the significant proportions between the binding functions $\texttt{asHomogeneousTransform}$, $\texttt{toMatlab}$, $\texttt{getWorldTransform}$ and the wrapper function $\texttt{getWorldTransform}$. 

The profilers have an uneven load on the process executing the Simulink model. The MATLAB interpreter profiler slows down the simulation by a factor of $1.48$ when monitoring the non optimised model, while having no impact on the model which uses Simulink functions. In the first case, it has many more functions to monitor in the \texttt{stepImpl} call tree, while in the second case it cannot evaluate the generated code from the Simulink blocks. This has to be accounted for when comparing the performance. 

\paragraph*{Selecting the reference block for comparing the performance} The Simulink profiler can only characterize the most outer block of the tested library, \texttt{RobotDynWithContacts}, and the Step MATLAB System block within, which computes the robot state evolution at each simulation step (\texttt{stepImpl} method). Prior to the framework optimisation, \texttt{stepImpl} runs all the required computations for producing the next robot state, while in the optimised framework, these computations are wrapped in the WBT library functions and so, run outside the Step MATLAB System block. This leaves us \texttt{RobotDynWithContacts} as the sole block suitable for observing the execution time improvement.
We consistently observe a very significant reduction of the execution time for each of the mentioned methods, by two orders of magnitude, as depicted in Figure \ref{fig:beforeAfterOptimComparison}.

\subsection{Controller Test and Execution Speed Performance}

For evaluating the simulator stability and execution real-time factor in the typical conditions of a controller performing a complex trajectory, we integrated a simulation system where a momentum-based whole-body torque controller\footnote{\url{https://github.com/robotology/whole-body-controllers/blob/v2.5.2/controllers/+floatingBaseBalancingTorqueControlWithSimulator}}~\cite{Nava_etal2016} controls a simulated $\textit{iCub}$\footnote{\url{http://www.icub.org/}} humanoid robot with 23 degrees of freedom, for balancing on a single foot while performing arms and free leg dynamic motions. For this purpose: we feed the controller torque setpoints to the simulator block $\texttt{RobotDynWithContacts}$ and close the loop using the simulator output; an IMU block was added for emulating the robot IMU sensor mounted on the base and generating measurements from the floating base state and linear acceleration; A friction model block was added before the simulator torque input to emulate the actuators dynamics disturbances; we ran the simulator, controller and visualizer at sampling rates of $\SI{1000}{\hertz}$, $\SI{100}{\hertz}$ and $\SI{20}{\hertz}$ respectively.

The simulation was run on a machine with an $\text{Intel}^{\circledR}$ $\text{Core}^{\text {TM }}$ i9 (8-core) CPU and 64(GB) RAM, on MATLAB R2020b, and was stable. For a simulated time of 45 seconds, without visualizer, the processing time measured through the Simulink profiler reached 46.5 seconds, resulting in a real-time factor of $0.97$.

This test can be reproduced by installing the simulator and controller packages, as well as their dependencies through a one-line installer or Conda package manager, as described in the project GitHub repository respective section\footnote{\url{https://github.com/ami-iit/matlab-whole-body-simulator#one-line-installation}}.

\subsection{Comparison with Gazebo}

\begin{figure}
    \centering
    \subfloat[]{
    \includegraphics[width=0.25\linewidth]{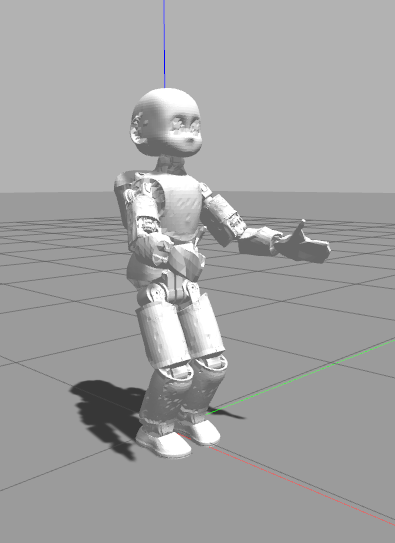}}
    \subfloat[]{
    \includegraphics[width=0.25\linewidth]{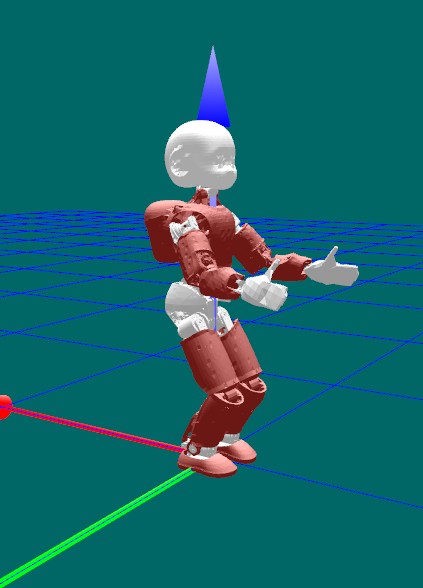}}
    \caption{The humanoid robot iCub in Gazebo and the proposed robot visualizer. (a) Gazebo. (b) Matlab/Simulink simulator}
    \label{fig:results:icub_in_gazebo_and_mwbs}
\end{figure}

\begin{figure}[!t]
    \centering
    \includegraphics[width=0.7\linewidth]{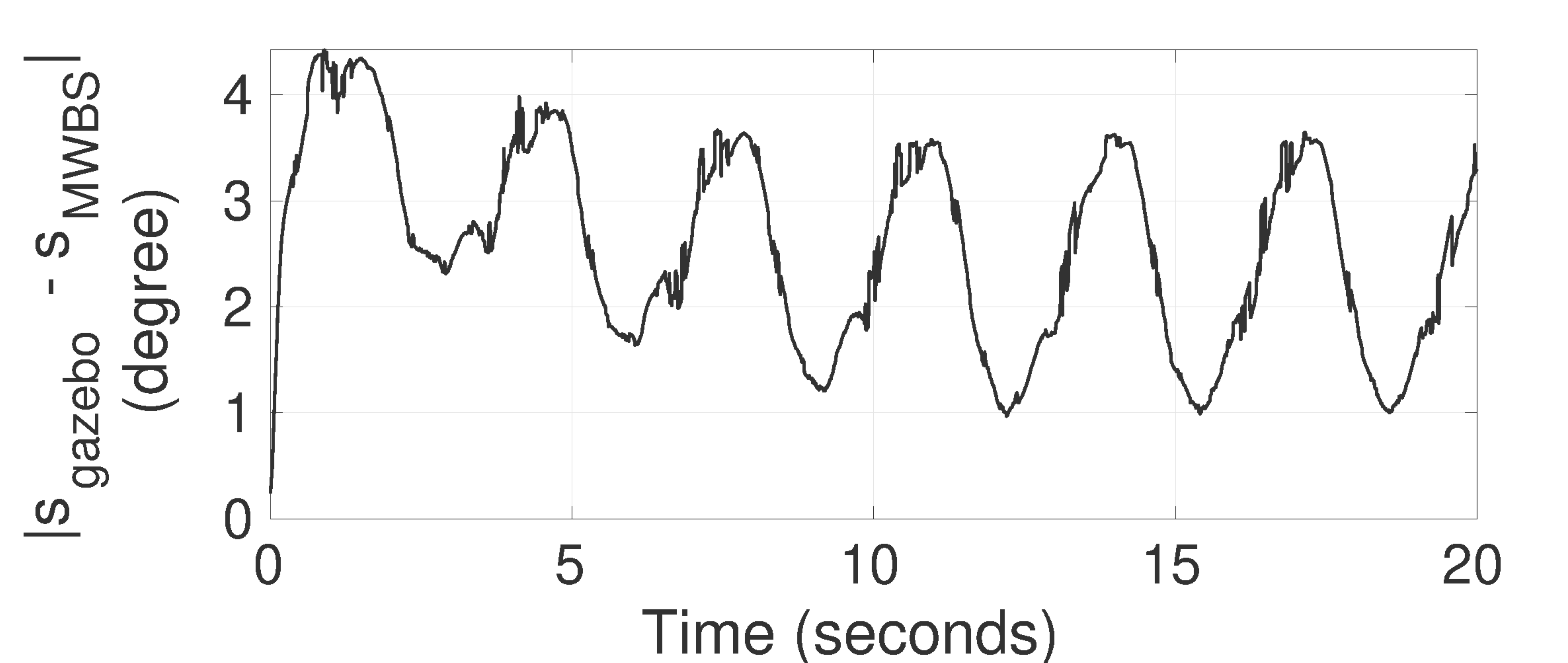}
    \caption{Comparison of robot joint position in the up and down simulation between Gazebo and the proposed simulator.}
    \label{fig:results:joint_pose_comparison_gazebo}
\end{figure}

Gazebo is a simulator widely used in robotics community~\cite{ivaldi2014tools}.
We compared Gazebo with our proposed simulator in a simulation scenario where the humanoid robot iCub performs an up and down action by bending and stretching the knees (see Figure \ref{fig:results:icub_in_gazebo_and_mwbs}).
For this purpose, we integrated the dynamic simulator with a PID controller with a gravity compensation term.
We tuned the PID controller for both Gazebo and our proposed simulator for the comparable joint position tracking (see Figure \ref{fig:results:joint_pose_comparison_gazebo}).

We run this simulation on a machine with $\text{Intel}^{\circledR}$ $\text{Core}^{\text {TM }}$ i7-10750H CPU and 32(GB) RAM, on MATLAB R2020b.
We recorded the execution times for the simulation with Gazebo and the Matlab/Simulink simulator through the Simulink profiler.
The integration step time for the core dynamics is $\SI{1000}{\hertz}$ and the step time for the robot visualizer is $\SI{10}{\hertz}$.
For performing 20 seconds simulation, the execution times for the simulation with Gazebo and our proposed simulator are about 23 and 32 seconds respectively.
Therefore, the real-time factors are respectively 0.87 and 0.63.
For the sake of simulation speed, in our proposed simulator, the simulator could be used with lower visualization step time or the dynamics core could be used without the robot visualizer.
In this way, the simulation execution time decreases to 18 seconds, resulting in a real-time factor of 1.11 for the up and down simulation scenario.

\subsection{Simulation of a robot with spherical feet}

In this section we illustrate the flexibility of our proposed simulation framework in integrating new features, considering as an example the modelling and simulation of a robot with spherical feet.

The proposed simulator is originally developed for free-floating base robots with rectangular feet.
For extending such a simulator for a robot with spherical feet, we need to revise the contact and impact models that represent the robot-ground interaction.
In the simulator implementation, the contact and impact models are summarized in the MATLAB class \texttt{Contacts} that could be simply revised.

For a rectangular foot, the distributed reaction forces applied to the foot are considered by pure forces applied to the four end tips of the feet, called contact vertices.
For spherical feet, instead, the reaction forces are pure forces applied to the lowest point of the feet.
Thus, the spherical foot, in contrast to a rectangular foot, has 1 time-varying contact vertex.
Therefore, for extending the contact/impact models for a spherical foot, one needs two features, namely the capability to define a desired number of contact vertices and the capability to define time-varying contact vertices.

For providing such capabilities: (1) we change the hard-coded value denoting the number of contact vertices to an input from the user; (2) we add some functions to the class \texttt{Contact} for the computation of contact point features like position or Jacobian; (3) we define a new input denoting the feet shape for the class \texttt{Contact}; (4) we add in the class \texttt{Contact} the logic selecting the proper functions which compute the contact vertices features relative to the feet shape.

Using the above revisions, one could simulate a robot with spherical feet.
\section{Conclusion}\label{sec:conclusion}

In this work, we have presented an open-source MATLAB/Simulink simulator for rigid body systems interacting with the ground. It provides an easy-to-use interface for plugging-in and parameterizing manipulators, free-floating robots, e.g.: humanoid robots; robot models with open-chain or closed-chain kinematics; robots with rectangular or spherical feet.

Beyond these features, the initial motivation for developing such a simulator was to improve the flexibility in the implementation of dynamics computations algorithms or physics models while keeping acceptable simulation performances compared to reference available simulators like Gazebo.
We achieved that goal by using object-oriented MATLAB features, like classes, and MEX-based bindings to C++ dynamics algorithms in the iDynTree external library.
This approach turned to be very costly for the execution speed performance, as shown in the MATLAB/Simulink profiling in Section~\ref{sec:results}. We tackled the problem by dropping pure bindings, which allowed code generation and replacing them with programmatic calls to native or custom Block Factory based Simulink blocks. This combined block-based and programmatic approach improved the design flexibility while matching comparable execution speed performance met with a reference simulator like Gazebo.

Thus, the benefit of the proposed library lies in adding to the intrinsic Simulink rapid prototyping, visual data analysis and profiling features, the possibility of: quickly editing and adding sensors and motor dynamics, growing away from simplified to more complex actuator and friction models; changing seamlessly simulation requirements and model parameters through Dynamic Mask Subsystems and bus data exchange.

\bibliographystyle{IEEEtran}
\bibliography{main}

\end{document}